%% file: main.tex
\pdfoutput=1
\documentclass[11pt]{article}

\usepackage[]{acl}

\usepackage{times}
\usepackage{latexsym}
\usepackage{amsmath}
\usepackage{amsfonts}
\usepackage{graphicx}
\usepackage{subfig}
\usepackage{color}
\usepackage{arydshln}
\usepackage{multirow}
\usepackage{booktabs} 
\usepackage{enumitem}
\usepackage{hyperref}

\usepackage[T1]{fontenc}
\usepackage[utf8]{inputenc}

\usepackage{microtype}

\usepackage{inconsolata}

\usepackage{algorithm}
\usepackage{algpseudocode}

\ifx\figurename\undefined \def\figurename{Figure}\fi
\ifx\figurenames\undefined \def\figurenames{Figures}\fi
\ifx\tablename\undefined \def\tablename{Table}\fi
\ifx\tablenames\undefined \def\tablenames{Tables}\fi
\ifx\equationname\undefined \def\equationname{Equation}\fi
\ifx\equationnames\undefined \def\equationnames{Equations}\fi

\title{iACOS: Advancing Implicit Sentiment Extraction with Informative and Adaptive Negative Examples}

\author{Xiancai Xu\footnotemark[1] \and Jia-Dong Zhang\footnotemark[1]~~\footnotemark[2] \and Lei Xiong \and Zhishang Liu\\
         Brands \& Consumers Research Institute, Enbrands Inc., Shenzhen, China \\ 
         \{essen, zhangjd.1, xiongl.1, liuzs.1\}@enbrands.com}
\begin{document}
\maketitle
\renewcommand{\thefootnote}{\fnsymbol{footnote}}
\footnotetext[1]{Equal contribution, alphabetical order of surnames.}
\footnotetext[2]{Corresponding author.}

\begin{abstract}
Aspect-based sentiment analysis (ABSA) have been extensively studied, but little light has been shed on the quadruple extraction consisting of four fundamental elements: aspects, categories, opinions and sentiments, especially with implicit aspects and opinions. In this paper, we propose a new method \textbf{iACOS} for extracting \textbf{I}mplicit \textbf{A}spects with \textbf{C}ategories and \textbf{O}pinions with \textbf{S}entiments. \textbf{First}, iACOS appends two implicit tokens at the end of a text to capture the context-aware representation of all tokens including implicit aspects and opinions. \textbf{Second}, iACOS develops a sequence labeling model over the context-aware token representation to co-extract explicit and implicit aspects and opinions. \textbf{Third}, iACOS devises a multi-label classifier with a specialized multi-head attention for discovering aspect-opinion pairs and predicting their categories and sentiments simultaneously. \textbf{Fourth}, iACOS leverages informative and adaptive negative examples to jointly train the multi-label classifier and the other two classifiers on categories and sentiments by multi-task learning. \textbf{Finally}, the experimental results show that iACOS significantly outperforms other quadruple extraction baselines according to the F1 score on two public benchmark datasets.
\end{abstract}
\renewcommand{\thefootnote}{\arabic{footnote}}
\input{introduction_naacl}

\input{related_work}

\input{modeling_naacl}

\input{experiments_naacl}
\input{conclusion_naacl}

%\section*{Limitations}
%EMNLP 2023 requires all submissions to have a section titled ``Limitations'', for discussing the limitations of the paper as a complement to the discussion of strengths in the main text. This section should occur after the conclusion, but before the references. It will not count towards the page limit.  

\bibliography{allbibliography}
\bibliographystyle{acl_natbib}

%\appendix
%\section{Example Appendix}
%\label{sec:appendix}

\end{document}

%% file: introduction_naacl.tex
\section{Introduction}
Aspect-based sentiment analysis (ABSA) has gained continuous attention during the last decade due to its broad application~\cite{Pontiki_Galanis_Pavlopoulos_2014,Pontiki_Galanis_Papageorgiou_2015,Pontiki_Galanis_Papageorgiou_2016}. ABSA aims to extract tuples consisting of closely related elements including the \textit{aspect term}, \textit{opinion term}, \textit{aspect category} and \textit{sentiment polarity}. The aspect term refers to a specific word or phrase in a text that is being evaluated, while the opinion term is a subjective statement in the text that expresses a personal sentiment on the aspect term. Both the aspect term and opinion term are typically classified into a predefined category and sentiment polarity, respectively. 
%There are various ABSA tasks that can be generally classified into six groups in terms of the extracted elements, namely, the aspect-opinion pair extraction~\cite{Chen_Liu_Wang_2020,Zhao_Huang_Zhang_2020,Gao_Wang_Liu_2021}, aspect-sentiment pair extraction~\cite{He_Lee_Ng_2019,Li_Bing_Li_2019,Chen_Qian_2020}, category-sentiment pair extraction~\cite{Ma_Peng_Cambria_2018,Hu_Zhao_Zhang_2019,Cai_Tu_Zhou_2020,Dai_Peng_Chen_2020}, aspect-opinion-sentiment triplet extraction~\cite{Peng_Xu_Bing_2020,Wu_Ying_Zhao_2020,Xu_Li_Lu_2020,Mao_Shen_Yu_2021,Xu_Chia_Bing_2021,Yan_Dai_Ji_2021,Chen_Chen_Sun_2022}, aspect-category-sentiment triplet extraction~\cite{Wan_Yang_Du_2020}, and aspect-category-opinion-sentiment quadruple extraction~\cite{Cai_Xia_Yu_2021,Zhang_Deng_Li_2021,Bao_Zhongqing_Jiang_2022,Mao_Shen_Yang_2022,Xiong_Yan_Wu_2023}. 
Most of the existing works only extract explicit aspects and opinions but completely ignore the implicit ones that are absent from texts. Some works consider the extraction of implicit aspects~\cite{Cai_Tu_Zhou_2020,Wan_Yang_Du_2020,Zhang_Li_Deng_2021,Zhang_Deng_Li_2021,Mao_Shen_Yang_2022}, implicit opinions~\cite{Setiowati_Djunaidy_Siahaan_2022}, or both~\cite{Cai_Xia_Yu_2021,Peper_Wang_2022,Xiong_Yan_Wu_2023,Bao_Jiang_Wang_2023,Bao_Wang_Zhou_2023,Hu_Bai_Wu_2023}. 

In particular, the study~\cite{Cai_Xia_Yu_2021} firstly attempts to extract implicit aspects and opinions simultaneously, because real textual reviews often contain a significant amount of implicit aspects and opinions.
For example, in the product review \textit{``Looks nice and the surface is smooth, but certain apps take seconds to respond''}~\cite{Cai_Xia_Yu_2021}, ``surface'' is an aspect term and classified into the ``Design'' category, ``smooth'' is the opinion term toward this aspect with the ``Positive'' sentiment. The four elements constitute an quadruple ``surface-Design-smooth-Positive''. Obviously, there are two more quadruples: ``null-Design-nice-Positive'' and ``apps-Software-null-Negative'', where ``null'' stands for an implicit aspect or opinion term that does not appear in the given text. 
Recently, the two studies~\cite{Peper_Wang_2022,Xiong_Yan_Wu_2023} have improved implicit quadruple extraction based on contrastive learning, a method that constructs positive and negative examples for each anchor (i.e., a training example). This method attempts to minimize the distance between the anchor and positive examples and maximize the distance between the anchor and negative examples in the latent representation space. Unfortunately, \textbf{most existing studies suffer from two limitations}: (1)~\textbf{Uninformative negative examples.} In contrastive learning, it is crucial to sample informative negative examples that are difficult to distinguish from the positive examples~\cite{Schroff_Kalenichenko_Philbin_2015}. However, existing studies often fail to generate such informative negative examples due to the intrinsic nature of random perturbation methods. (2)~\textbf{Non-adaptive negative examples.} The negative examples lack adaptiveness, as their sampling is not influenced by the current model parameters~\cite{Daghaghi_Medini_Meisburger_2021}. As a result, there is significant scope to enhance performance in extracting aspect-category-opinion-sentiment quadruples from texts.

Therefore, this paper proposes a new method based on informative and adaptive negative examples, namely \textbf{iACOS}, for extracting \textbf{I}mplicit \textbf{A}spects with \textbf{C}ategories and \textbf{O}pinions with \textbf{S}entiments. \textbf{First}, iACOS employs the pre-trained encoder BERT~\cite{Devlin_Chang_Lee_2018} to get the context-aware token representation of a text, by which a large amount of knowledge contained in BERT can be transferred into iACOS. Meanwhile, iACOS appends two implicit tokens at the end of texts to capture the semantic representation of implicit aspects and opinions, respectively. \textbf{Second}, iACOS builds a sequence labeling model over the context-aware token representation by extending the BIOES\footnote{BIOES is a tagging scheme for sequence labeling and BIOES denotes Begin, Inside, Outside, End and Single, respectively.} tagging scheme to co-extract explicit and implicit aspects and opinions; the aspects-opinion co-extraction is preferentially executed, since it is a relatively simple task~\cite{Wang_Pan_Dahlmeier_2017,Wang_Pan_2018,Dai_Song_2019} and our extended sequence labeling model can accurately generate aspect and opinion candidates for other subsequent tasks. \textbf{Third}, iACOS develops a multi-label classifier with a specialized multi-head attention to predict the category-sentiment combination label of each aspect-opinion candidate pair; this classifier is an end-to-end method for discovering aspect-opinion pairs and predicting their categories and sentiments at the same time to alleviate error propagation in the pipeline solution~\cite{Peng_Xu_Bing_2020,Cai_Tu_Zhou_2020}. \textbf{Fourth}, iACOS constructs negative examples based on the aspect-opinion co-extraction results to train the classifier. These negative examples are informative and adaptive to current model parameters due to two reasons. (1)~They are carefully selected or constructed examples that closely resemble positive examples but are actually negative. Therefore, these examples help in refining the model's ability to distinguish subtle differences between aspects, opinions, categories, and sentiments that are similar yet distinct. The informative nature of these examples stems from their relevance and challenge to the model, pushing it to learn more nuanced differentiations. (2)~These examples are dynamically generated based on the current state of the model during training. Unlike static negative examples used in traditional models, adaptive examples evolve as the model learns, ensuring that the model is consistently challenged. This adaptiveness is critical in iACOS, as it allows the model to improve its understanding of complex sentiment relationships continuously. Additionally, the negative examples are used to jointly train two other classifiers: one for predicting aspect categories and another for opinion sentiments, using a multi-task learning approach. In this study, we address the critical shortage of labeled data impeding complex ABSA tasks by augmenting training data with negative examples, rather than employing contrastive learning.

The main contributions are listed below:
\begin{itemize}[leftmargin=11pt]
	\item We propose a new method iACOS for extracting aspect-category-opinion-sentiment quadruples. iACOS unifies the extraction of \textit{explicit} and \textit{implicit} aspects and opinions based on a sequence labeling model. We develop a multi-label classifier for integrating the prediction on categories, sentiments, and their matched pairs into one unified task to alleviate error propagation in the pipeline solution.
	
	\item We leverage informative and adaptive negative examples for jointly training multiple tasks, which significantly improves the effectiveness of quadruple extraction. To the best of our knowledge, this is the first attempt to construct informative and adaptive negative examples as data augmentation for ABSA tasks.
	
	\item We conduct extensive experiments to verify the effectiveness of iACOS on the two public benchmark datasets~\cite{Cai_Xia_Yu_2021} for quadruple extraction. Experimental results show that iACOS improves the F1 score significantly, in comparison to other state-of-the-art quadruple extraction techniques for implicit aspects and opinions. Our source code is publicly released at \href{https://github.com/jiadongzh/iacos}{https://github.com/jiadongzh/iacos}.
\end{itemize}

%The rest of this paper is organized as follows. We first highlight related work and then present iACOS, followed by experimental evaluation. Finally, We concludes this paper.

The rest of this paper is organized as follows. We highlight related work in Section~\ref{sec:rw}. Our iACOS is presented in Section~\ref{sec:modeling}, followed by experimental evaluation in Section~\ref{sec:exp}. Finally, Section~\ref{sec:conclusion} concludes this paper.

%% file: related_work.tex
\section{Related Work}\label{sec:rw}
This section reviews the recent advances of sentiment quadruple extraction and contrastive learning.

\textbf{Quadruple Extraction.} Recently, \citet{Cai_Xia_Yu_2021} introduce a new task, named aspect-category-opinion-sentiment quadruple extraction, construct two new datasets for this new task, and benchmark the task with four baseline systems. Later, most works~\cite{Zhang_Deng_Li_2021,Bao_Zhongqing_Jiang_2022,Mao_Shen_Yang_2022,Bao_Jiang_Wang_2023,Bao_Wang_Zhou_2023,Hu_Bai_Wu_2023} apply the sequence-to-sequence model T5 to generate a list of quadruples for a given sentence and can be differentiated from one another in terms of the formats of quadruples. For instance, the literature~\cite{Zhang_Deng_Li_2021} represents each quadruple as a paraphrase sentence, the reference~\cite{Bao_Zhongqing_Jiang_2022,Bao_Jiang_Wang_2023,Bao_Wang_Zhou_2023} formats all quadruples as an opinion tree with linearized order, and the research~\cite{Mao_Shen_Yang_2022} denotes each quadruple as an independent path of a tree without linearized order. Other studies~\cite{Gao_Fang_Liu_2022,Wang_Xia_Yu_2022,Varia_Wang_Halder_2022} develop a unified generative framework based on the T5 model with instructional prompts for a variety of ABSA tasks including quadruple extraction.

\textbf{Implicit Aspects and Opinions.} Although there are so many existing works on extracting aspects and opinions, most of them completely ignore the implicit aspects and opinions that do not appear in texts. Some recent studies pay attention on extracting implicit aspects~\cite{Cai_Tu_Zhou_2020,Wan_Yang_Du_2020,Zhang_Li_Deng_2021,Zhang_Deng_Li_2021,Mao_Shen_Yang_2022}. For example, the study~\cite{Cai_Tu_Zhou_2020} does not mine implicit aspects but directly derives their corresponding categories, the research~\cite{Wan_Yang_Du_2020} handles implicit aspects by classifying whether aspect terms exist in the sentence for a given category-sentiment pair, and other works~\cite{Zhang_Li_Deng_2021,Zhang_Deng_Li_2021,Mao_Shen_Yang_2022} naturally represent implicit aspect terms as ``null'' in the output sequence based on the sequence-to-sequence model T5. In contrast, few work considers the extraction of implicit opinions. For instance, the work~\cite{Setiowati_Djunaidy_Siahaan_2022} infers implicit opinions via learning a co-occurrence matrix between aspects and opinions. More comprehensively, the study~\cite{Cai_Xia_Yu_2021} is the first to manage implicit aspects and opinions simultaneously by predicting whether the implicit aspect or opinion exists in a given text. Later, \citet{Bao_Jiang_Wang_2023,Bao_Wang_Zhou_2023} insert two fake tokens at the beginning of a sentence as the implicit aspect and opinion term; other works~\cite{Peper_Wang_2022,Xiong_Yan_Wu_2023,Hu_Bai_Wu_2023} naturally denote implicit aspect terms as ``it'' or ``null'' and implicit option terms as ```null''.

\textbf{Contrastive Learning.} The works~\cite{Peper_Wang_2022,Xiong_Yan_Wu_2023} enhance implicit quadruple extraction through contrastively learning a sequence-to-sequence model. The contrastive learning utilizes positive and negative samples to produce better input representations by pushing a given anchor with its positive sample closer together while pulling the anchor with its negative sample farther apart in the latent space. It is essential to select negative samples that are challenging to differentiate from positive ones for effective contrastive learning~\cite{Schroff_Kalenichenko_Philbin_2015}. Specifically, the work~\cite{Peper_Wang_2022} perturbs each anchor representation with a random dropout probability to obtain positive and negative samples, while the work~\cite{Xiong_Yan_Wu_2023} constructs negative samples by randomly replacing aspect and opinion words in positive samples. Nevertheless, existing works often fall short in producing informative and adaptive negative samples, owing to the inherent limitations of random techniques that are independent of current model parameters~\cite{Daghaghi_Medini_Meisburger_2021}. In this study, we concentrate on generating more informative and adaptive samples for data augmentation instead of contrastive learning, due to a lack of labeled training data.

%% file: modeling_naacl.tex
\section{The Proposed iACOS}\label{sec:modeling}
%We first define the research problem in Section~\ref{subsec:ps}, then introduce the inference process in Sections~\ref{subsec:co-ex} and~\ref{subsec:quad-ex} (see the left box of \figurename~\ref{fig:framework}), and finally present the training process in Sections~\ref{subsec:ns} and~\ref{subsec:mtl} (see the right box of \figurename~\ref{fig:framework}).

%We first define the research problem, then introduce the inference process (see the left box of \figurename~\ref{fig:framework}), and finally present the training process (see the right box of \figurename~\ref{fig:framework}).

We define the research problem in Section~\ref{subsec:ps}, introduce the inference process in Sections~\ref{subsec:co-ex}-\ref{subsec:quad-ex}, and present the training process in Sections~\ref{subsec:ns}-\ref{subsec:mtl}.

\begin{figure*}[!t]
	\centering
	\includegraphics[width=0.9\textwidth]{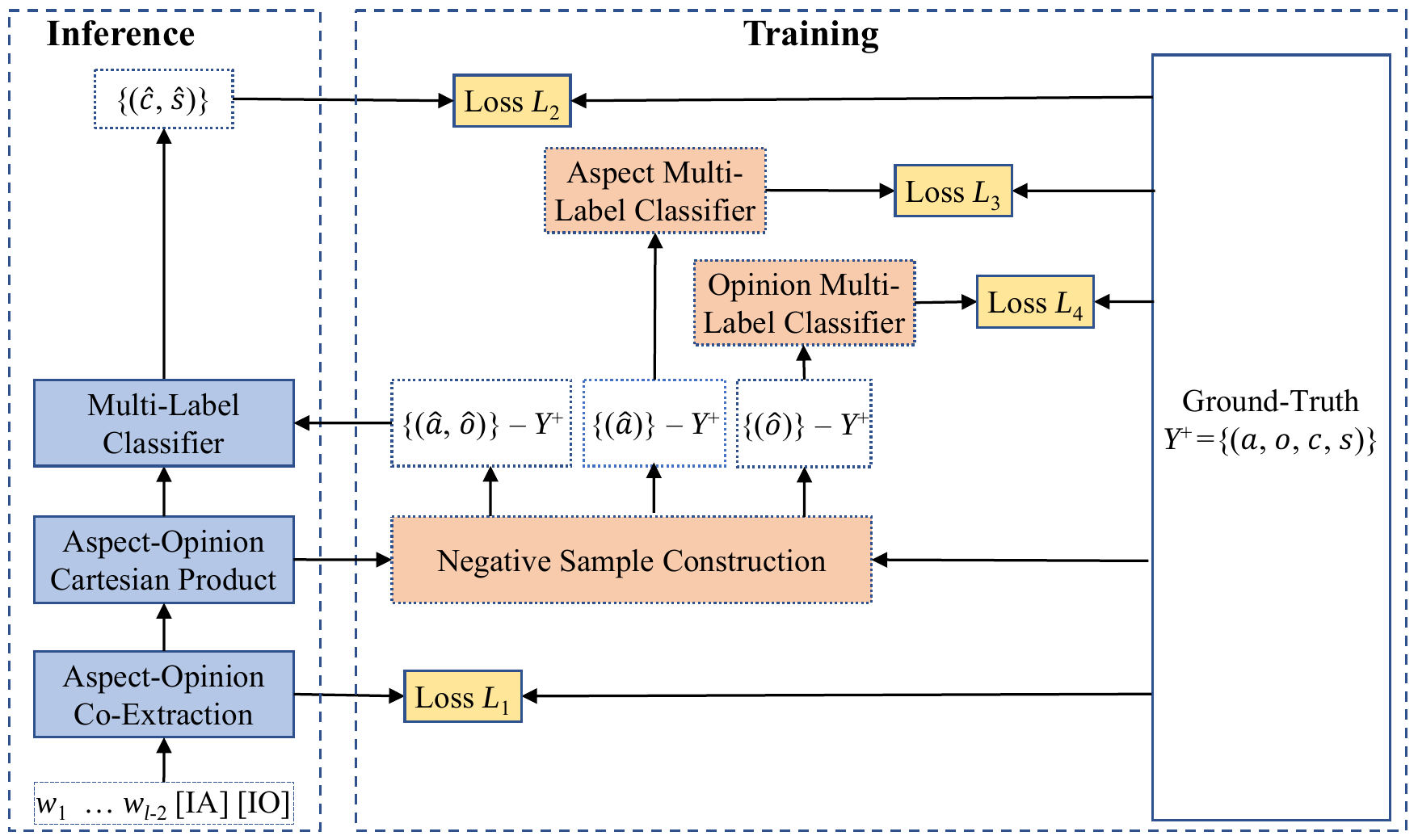}
	\caption{Framework of iACOS: left box for inference and right box for training with multi-tasking learning, in which negative sample construction is an important module.}
	\label{fig:framework}
\end{figure*}

\subsection{Problem Statement}\label{subsec:ps}

We first define basic concepts and the research problem for this paper.

\textbf{Token.} A text, e.g., a product review, is often segmented into a sequence of words or tokens $\left<w_1, w_2, \dots, w_l\right>$. Both words and tokens are used interchangeably in this paper.
	
\textbf{Aspect term.} An aspect term $a$ refers to a word span $\left<w_j, \dots, w_{j+m}\right>$ ($1 \le j \le j+m \le l$) in the text that represents an attribute or feature being evaluated by the corresponding opinion term(s). All aspects in the text constitute a set $A$.
	
\textbf{Opinion term.} An opinion term $o$ refers to a word span $\left<w_k, \dots, w_{k+n}\right>$ ($1 \le k \le k+n \le l$) in the text that expresses a personal sentiment on the corresponding aspect term(s).  All opinions in the text constitute a set $O$.
	
\textbf{Category.} A category $c \in C$ is a predefined label that is used to classify an aspect $a$.

\textbf{Sentiment.} A sentiment polarity, or simply sentiment $s \in S$, represents a predefined semantic orientation (e.g., positive, negative, or neutral) expressed by an opinion $o$.

\textbf{Quadruple.} A quadruple $(a, o, c, s)$ represents the correlation among its four elements.

\textbf{Research problem.} Given a set of training texts, each containing $l$ words $\left<w_1, w_2, \dots, w_l\right>$ with ground-truth quadruples $Y^+ = \{(a, o, c, s)\}$, we aim to learn a model to extract a set of quadruples $\{(\hat a, \hat o, \hat c, \hat s)\}$ from a new text. 

%where $a$ is an aspect term and denoted as word span $\left<w_j, \dots, w_{j+m}\right>$, $o$ is an opinion term and denoted as word span $\left<w_k, \dots, w_{k+n}\right>$, and the aspect and opinion terms are classified into a category $c \in C$ and sentiment $s \in S$, respectively

\subsection{Explicit and Implicit Aspect-Opinion Co-Extraction}\label{subsec:co-ex}

\textbf{Representation of implicit aspects and opinions.} People often do not explicitly express their opinions on aspects; it is common to observe implicit aspects and opinions which are absent from a given text. To handle these implicit aspects and opinions, iACOS designs two implicit tokens to capture their semantic representation as done for explicit tokens. As depicted in the left box of \figurename~\ref{fig:framework}, at first the two specialized tokens ``[IA]'' and ``[IO]'' are appended at the end of a given text. Then iACOS feeds the appended text into the pre-trained encoder BERT~\cite{Devlin_Chang_Lee_2018} to learn the context-aware representation for all tokens, denoted as:
\begin{multline}
\left<{\bf h}_1, \dots, {\bf h}_{l-2}, {\bf h}_\text{[IA]}, {\bf h}_\text{[IO]}\right> = \\ BERT(\left<w_1, \dots, w_{l-2}, \text{[IA]}, \text{[IO]}\right>),
\end{multline}
where without loss of generality, the last two tokens $w_{l-1}=$ [IA] and $w_l=$ [IO] denote implicit aspects and opinions, respectively, and ${\bf h}$ is the context-aware representation of a token. It is worth emphasizing  that ${\bf h}_\text{[IA]}$ and ${\bf h}_\text{[IO]}$ are the semantic representations of implicit aspects and opinions which are learned from the whole text.

\textbf{Aspects-opinion co-extraction.} As we can see, obtaining implicit aspects and opinions is easy with the two specialized tokens [IA] and [IO]. However, a model is still required to co-extract explicit aspects and opinions. To this tend, iACOS builds a sequence labeling model with the extended BIOES tagging scheme over the context-aware token representation. In particular, the extended BIOES tagging scheme consists
of nine tags: $T = \text{\{B-A, I-A, E-A, S-A, B-O, I-O, E-O, S-O, O\}}$ with the suffix indicating the tag for aspects or opinions. Formally, we can predict the probability distribution ${\bf p}_i \in \mathbb{R}^9$ of each token ${\bf h}_i \in \mathbb{R}^\mathtt{d}$ over nine tags via a linear layer with Softmax:
\begin{equation}\label{eq:bio-prob}
{\bf p}_i = Softmax({\bf W}_1{\bf h}_i+{\bf b}_1),
\end{equation}
where ${\bf W}_1 \in \mathbb{R}^{9 \times \mathtt{d}}$ and ${\bf b}_1 \in \mathbb{R}^9$ are the weight matrix and bias vector, respectively. The tag is
\begin{equation}\label{eq:bio-predict}
	\hat y_i = \arg \max_{t \in T} p_{i,t}\text{~and~}p_{i,t} \in {\bf p}_i.
\end{equation}
The predicted $\hat y_i$ for a given text can be easily decoded to a set of aspects denoted as $A = \{\hat a\}$ and a set of opinions denoted as $O = \{\hat o\}$. It is worth noting that the tokens [IA] and [IO] for implicit aspects and opinions are always added into $A$ and $O$, respectively. From now on, it is unified to process both \textit{explicit} and \textit{implicit} aspects and opinions.  

\subsection{Multi-label Classifier with Multi-head Attention for Quadruple Extraction}\label{subsec:quad-ex}

Given aspects $A$ and opinions $O$ from \equationautorefname~(\ref{eq:bio-predict}), iACOS simultaneously performs category and sentiment prediction, and their pair matching process to alleviate the error propagation of the pipeline solution. iACOS considers the Cartesian product $C \times S$ as the combination label set, and predicts multiple combination labels for each aspect-opinion pair because implicit aspects and opinion may have multiple category and sentiment labels, respectively. Given a pair of aspect $\hat a = \left<w_j, \dots, w_{j+m}\right> \in A$ and opinion $\hat o=\left<w_k, \dots, w_{k+n}\right> \in O$, iACOS concatenates the vectors of all the tokens in the aspect and opinion by
\begin{multline}\label{eq:pair-emb}
	{\bf H}^{(\hat a \hat o)} = [{\bf h}_j, \dots, {\bf h}_{j+m}, \\ {\bf h}_k, \dots, {\bf h}_{k+n}] \in \mathbb{R}^{(m+n+2) \times \mathtt{d}},
\end{multline}
and exploits a multi-head attention~\cite{Vaswani_Shazeer_Parmar_2017} over them to get the attention vector
\begin{equation}\label{eq:pair-att}
	{\bf h}^{(\hat a \hat o)} =  MultiHead({\bf q}, {\bf H}^{(\hat a \hat o)}, {\bf H}^{(\hat a \hat o)}) \in \mathbb{R}^\mathtt{d},
\end{equation}
where ${\bf q} \in \mathbb{R}^\mathtt{d}$ is the trainable query, ${\bf H}^{(\hat a \hat o)}$ are the keys and values, and the head number is set to 8 by default. Further, the attention vector is fed into a linear layer with Sigmoid to obtain the probability of every combination label: 
\begin{equation}\label{eq:pair-prob}
	{\bf p}^{(\hat a \hat o)} = Sigmoid({\bf W}_2{\bf h}^{(\hat a \hat o)} + {\bf b}_2) \in \mathbb{R}^{|C \times S|},
\end{equation}
in which ${\bf W}_2 \in \mathbb{R}^{|C \times S| \times \mathtt{d}}$ and ${\bf b}_2 \in \mathbb{R}^{|C \times S|}$ are the weight matrix and bias vector, respectively. Each entry in ${\bf p}^{(\hat a \hat o)}$ from \equationautorefname~(\ref{eq:pair-prob}) with the probability larger than 0.5 indicates the corresponding category $c$ and sentiment $s$, i.e., one predicted quadruple $(\hat a, \hat o, \hat c, \hat s)$.

\subsection{Constructing Informative and Adaptive Negative Samples}\label{subsec:ns}
\textbf{Optimization objectives.} As depicted in the right box of \figurename~\ref{fig:framework}, without loss of generality, considering a text with words $\left<w_1, w_2, \dots, w_l\right>$ and ground-truth quadruples $Y^+ = \{(a, o, c, s)\}$, it is easy to obtain the BIOES tags of all tokens in terms of both ground-truth aspects $\{a\}$ and opinions $\{o\}$. Therefore, we can learn ${\bf W}_1$ and ${\bf b}_1$ in \equationautorefname~(\ref{eq:bio-prob}) through minimizing the cross-entropy loss:
\begin{equation}\label{eq:loss-bio}
	L_1 = \frac{1}{l} \sum\nolimits_{i=1}^{l} {\bf y}_i \cdot \log{\bf p}_i,
\end{equation}
where ${\bf y}_i \in \mathbb{R}^9$ is the one-hot vector of the ground-truth tag of token $w_i$. Moreover, $Y^+ = \{(a, o, c, s)\}$ may contain quadruples with equal $(a, o)$ pair but different combinations  $(c, s) \in C \times S$, that is, an aspect-opinion pair may have multiple combination labels. For instance, in the product review \textit{``so what you really end up paying for is the restaurant not the food''}, the two quadruples ``restaurant-Price-null-Negative'' and ``restaurant-Ambience-null-Neutral'' suggest the restaurant-[IO] pair has two combination labels (Price, Negative) and (Ambience, Neutral)~\cite{Cai_Xia_Yu_2021}. Hence, $Y^+ = \{(a, o, c, s)\}$ can be reduced to  $Y^+ = \{(a, o, \{(c, s)\})\} = \{(a, o, {\bf y}^{(ao)})\}$ in which ${\bf y}^{(ao)}$ is the multiple ground-truth combination labels. Accordingly, ${\bf W}_2$ and ${\bf b}_2$ in \equationautorefname~(\ref{eq:pair-prob}) can be learned through minimizing the binary cross-entropy loss:
\begin{multline}\label{eq:loss-pair}
	L_2^+ = \frac{1}{|Y^+||C \times S|} \sum\nolimits_{(a, o, {\bf y}^{(ao)}) \in Y^+} [{\bf y}^{(ao)}\cdot \\ \log{\bf p}^{(ao)} + (1-{\bf y}^{(ao)})\cdot \log(1-{\bf p}^{(ao)})],
\end{multline}
where ${\bf p}^{(ao)}$ is the probability distribution on combination labels of aspect $a$ and opinion $o$, computed from \equationautorefname~(\ref{eq:pair-emb}) to \equationautorefname~(\ref{eq:pair-prob}). Unfortunately, when minimizing the loss $L_2^+$ in \equationautorefname~(\ref{eq:loss-pair}) with ground-truth quadruples $Y^+$, one problem is that $Y^+$ is often insufficient to learn a unified model for predicting categories, sentiments, and their matched pairs at the same time. 

\textbf{Negative sample construction.} To tackle this problem, iACOS exploits informative and adaptive negative samples to train the unified model. The negative samples are constructed based on the aspect-opinion co-extraction results and hard to be discriminated against ground-truth samples by the current unified model. Further, the method is adaptive since the negative samples are dependent on the input data and current dynamically updated parameters. The two characteristics are the key to acquire high-quality samples~\cite{Daghaghi_Medini_Meisburger_2021} and differentiate this method from existing static methods such as random sampling, frequency-based static sampling~\cite{Bengio_Senecal_2008,Mikolov_Sutskever_Chen_2013} or learning-based biased sampling~\cite{Bamler_Mandt_2020,Gutmann_Hyvarinen_2010}. Specifically, given the aspect-opinion co-extraction results from a text, i.e., the sets of predicted aspects $A = \{\hat a\}$ and opinions $O = \{\hat o\}$ according to \equationautorefname~(\ref{eq:bio-predict}), the Cartesian product $A \times O$ contains all $(\hat a, \hat o)$ pair candidates that are used to derive quadruples based on \equationautorefname~(\ref{eq:pair-prob}). In other words, the unified model must learn to tell these candidates apart: some pair candidates are present in the ground-truth quadruples $Y^+ = \{(a, o, {\bf y}^{(ao)})\}$ accompanied by their corresponding combination labels ${\bf y}^{(ao)}$, while others are not. To this end, iACOS subtracts the ground-truth quadruples $Y^+$ from the Cartesian product $A \times O$, simply denoted as $Y^- = A \times O - Y^+ = \{(\hat a, \hat o)\} - Y^+$ and considers the remainder pair candidates in $Y^-$ as negative samples. Accordingly, the binary cross-entropy loss is given by 
\begin{multline}\label{eq:loss-pair-neg}
	L_2^- = \frac{1}{|Y^-||C \times S|} \sum\nolimits_{(\hat a, \hat o, {\bf y}^{(\hat a \hat o)}) \in Y^-} [{\bf y}^{(\hat a \hat o)}\cdot \\ \log{\bf p}^{(\hat a \hat o)} + (1-{\bf y}^{(\hat a \hat o)})\cdot \log(1-{\bf p}^{(\hat a \hat o)})],
\end{multline}
where ${\bf y}^{(\hat a \hat o)} \in {\bf 0}^{|C \times S|}$ is a zero vector, i.e., the ground-truth label for the negative pair of aspect $\hat a$ and opinion $\hat o$. Finally, iACOS adds negative samples $Y^-$ into ground-truth quadruples $Y^+$ and then computes their loss together:
\begin{multline}\label{eq:loss-pair-total}
	L_2 = \frac{1}{|Y^+ \cup Y^-||C \times S|} \sum\nolimits_{(\tilde a, \tilde o, {\bf y}^{(\tilde a \tilde o)}) \in Y^+ \cup Y^-} \\ [{\bf y}^{(\tilde a \tilde o)}\cdot \log{\bf p}^{(\tilde a \tilde o)} + (1-{\bf y}^{(\tilde a \tilde o)})\cdot \log(1-{\bf p}^{(\tilde a \tilde o)})].
\end{multline}

\subsection{Multi-task Learning}\label{subsec:mtl}

It is straightforward to train the model for quadruple extraction based on minimizing the sum of losses $L_1$ and $L_2$. Due to the lack of training data, iACOS also leverages both the ground-truth quadruples $Y^+$ and negative samples $Y^-$ to jointly learn the other two related classifiers for predicting the category of aspects and sentiment of opinions, respectively. Similar to \equationautorefname~(\ref{eq:pair-emb}), iACOS separately concatenates the token vectors by 
\begin{equation}\label{cs-emb}
\begin{aligned}
&	{\bf H}^{(\hat a)} = [{\bf h}_j, \dots, {\bf h}_{j+m}] \in \mathbb{R}^{(m+1) \times \mathtt{d}}\text{~and~}\\ &	{\bf H}^{(\hat o)} = [{\bf h}_k, \dots, {\bf h}_{k+n}] \in \mathbb{R}^{(n+1) \times \mathtt{d}},
\end{aligned}
\end{equation}
and get the corresponding attention vectors
\begin{equation}\label{eq:cs-att}
\begin{aligned}
&	{\bf h}^{(\hat a)} =  MultiHead({\bf q}_1, {\bf H}^{(\hat a)}, {\bf H}^{(\hat a)})\text{~and~}\\
&	{\bf h}^{(\hat o)} =  MultiHead({\bf q}_2, {\bf H}^{(\hat o)}, {\bf H}^{(\hat o)}),
\end{aligned}
\end{equation}
which are fed into a linear layer with Sigmoid to obtain the probability distributions
\begin{equation}\label{eq:cs-prob}
\begin{aligned}
&	{\bf p}^{(\hat a)} = Sigmoid({\bf W}_3{\bf h}^{(\hat a)} + {\bf b}_3)  \in \mathbb{R}^{|C|}\text{~and~}\\
&	{\bf p}^{(\hat o)} = Sigmoid({\bf W}_4{\bf h}^{(\hat o)} + {\bf b}_4) \in \mathbb{R}^{|S|}.
\end{aligned}
\end{equation}
Further, similar to \equationautorefname~(\ref{eq:loss-pair-total}), iACOS minimizes the two binary cross-entropy losses
\begin{multline}\label{eq:loss-cat}
	L_3 = \frac{1}{|Y^+ \cup Y^-||C|} \sum\nolimits_{(\tilde a, {\bf y}^{(\tilde a)}) \in Y^+ \cup Y^-} [{\bf y}^{(\tilde a)}\cdot \\ \log{\bf p}^{(\tilde a)} + (1-{\bf y}^{(\tilde a)})\cdot \log(1-{\bf p}^{(\tilde a)})]\text{~and~}
\end{multline}
\begin{multline}\label{eq:loss-sent}
	L_4 = \frac{1}{|Y^+ \cup Y^-||S|} \sum\nolimits_{(\tilde o, {\bf y}^{(\tilde o)}) \in Y^+ \cup Y^-} [{\bf y}^{(\tilde o)}\cdot \\ \log{\bf p}^{(\tilde o)} + (1-{\bf y}^{(\tilde o)})\cdot \log(1-{\bf p}^{(\tilde o)})],
\end{multline}
where ${(\tilde a, {\bf y}^{(\tilde a)})}$ or ${(\tilde o, {\bf y}^{(\tilde o)})}$ denotes a projection of $Y^+ \cup Y^-$ on aspects or opinions for simplicity. Eventually, iACOS jointly trains all model parameters by minimizing the total loss with the Adam optimization algorithm on data batches: 
\begin{equation}\label{eq:loss-total}
	L = L_1 + L_2 + L_3 + L_4.
\end{equation}
The multi-task learning improves data efficiency and reduces overfitting because of shared context-aware representations ${\bf h}$ among these tasks.

%% file: experiments_naacl.tex
\section{Experiments}\label{sec:exp}
We present the evaluation setup in Section~\ref{subsec:es} and experimental results in Section~\ref{subsec:ra}.

%We present the experimental setup and experimental results in this section.

\begin{table}[htb]
	\footnotesize
	\centering
	\begin{tabular}{cccc}
		\toprule
		     &               & \textit{Restaurant} & \textit{Laptop}\\
		\midrule       
		     & \#Categories  & 13   & 121\\
		\midrule
			 &  \#Sentences  & 2,286 & 4,076\\
		\midrule
		                          & EA\&EO & 2,429 & 3,269\\
		\cmidrule{2-4}
		                          & IA\&EO & 530  & 910\\
		\cmidrule{2-4}
		     \#Quadruples              & EA\&IO & 350  & 1,237\\
		\cmidrule{2-4}
		                          & IA\&IO & 349  & 342\\
		\cmidrule{2-4}
		                          & All    & 3,658 & 5,758\\
		\bottomrule
	\end{tabular}
	\caption{Statistics of the two datasets from the work~\cite{Cai_Xia_Yu_2021}. E, I, A and O denote Explicit, Implicit, Aspect and Opinion, respectively.}
	\label{tab:data}
\end{table}

\begin{figure}[!t]
	\centering
	\subfloat[\textit{Restaurant}]{\label{subfig:rest16-convergent}%
		\includegraphics[width=0.5\columnwidth]{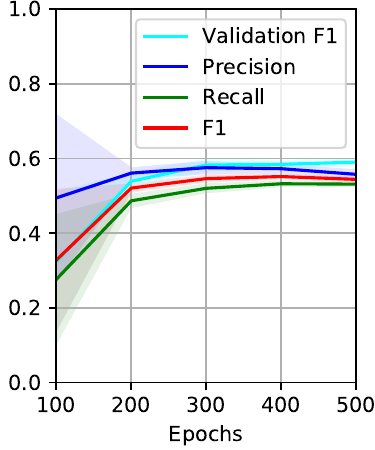}}
	\hfill
	\subfloat[\textit{Laptop}]{\label{subfig:laptop-convergent}%
		\includegraphics[width=0.5\columnwidth]{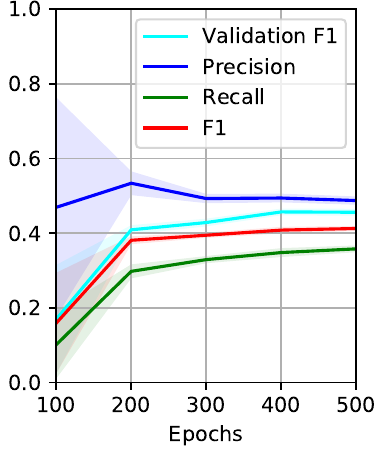}}
	\caption{Convergence analysis on iACOS}
	\label{fig:convergent}
\end{figure}

\subsection{Experimental Setup}\label{subsec:es}

\textbf{Datasets.}
We use two public benchmark datasets on the quadruple extraction task with implicit aspects and options from the work~\cite{Cai_Xia_Yu_2021} which reports the basic statistics of the two datasets in \tablename~\ref{tab:data}. We adopt exactly the same splits on the two datasets for training, validation and testing as the original work.

\textbf{Compared methods.}
We compare iACOS with the state-of-the-art baselines on quadruple extraction with implicit aspects and options listed below:
\begin{itemize}[leftmargin=11pt]
	%\item Double Propagation (DP): It is a double propagation and rule-based method for aspect-opinion-sentiment triple extraction~\cite{Qiu_Liu_Bu_2011}, and it has been adapted for the quadruple extraction task by first extracting all the aspect-opinion-sentiment triples, followed by assigning the aspect category for each extracted triple.
	
	%\item JET: It is an end-to-end framework for aspect-opinion-sentiment triple extraction~\cite{Xu_Li_Lu_2020}, and it has been adapted for the quadruple extraction task, similar to DP.
	
	\item TAS: It adapts the input transformation strategy of the target-aspect-sentiment model~\cite{Wan_Yang_Du_2020} to perform category-sentiment conditional aspect-opinion co-extraction, following by filtering out the invalid aspect-opinion pairs to form the final quadruples.
	
	\item Extract-Classify: It performs aspect-opinion co-extraction and predicts the sentiment polarity of the extracted aspect-opinion pair candidates conditioned on each category~\cite{Cai_Xia_Yu_2021}.
	
	\item {\sc Paraphrase}: It casts the quadruple extraction task to a paraphrase generation process that jointly detects all four elements~\cite{Zhang_Deng_Li_2021} and has been adapted for implicit aspects and opinions~\cite{Xiong_Yan_Wu_2023}.
	
	\item GEN-NAT-SCL: It uses a contrastive learning objective to aid quadruple prediction by encouraging the model to produce input representations~\cite{Peper_Wang_2022}.
	
	\item BART-CRN: It is a BART-based contrastive and retrospective network (BART-CRN) that learns the associations among all types of quadruples~\cite{Xiong_Yan_Wu_2023}.
\end{itemize}

To ensure fairness, we focus on models with BERT or BART backbones, excluding the larger and stronger T5 models~\cite{Bao_Jiang_Wang_2023,Bao_Wang_Zhou_2023,Hu_Bai_Wu_2023}. Our work centers on a novel method of employing informative and adaptive negative examples for joint multi-task training, which could improve performance when applied to stronger backbones like T5. Thus, our approach could lead to surpassing current top results.

\textbf{Evaluation metrics.}
In line with existing studies~\cite{Zhang_Deng_Li_2021,Cai_Xia_Yu_2021}, the Precision, Recall, and F1 score are adopted as the main evaluation metrics. Moreover, we view a predicted quadruple as correct if and only if the four elements as well as their combination are exactly the same as those in the ground-truth quadruples.

\textbf{Experimental settings.}
We adopt the pre-trained BERT as the backbone and use the AdamW optimizer to minimize the total loss. The hyper-parameters are determined based on existing studies~\cite{Zhang_Deng_Li_2021,Cai_Xia_Yu_2021} and several trials on the validation data instead of exhausting grid search. By default, we respectively set the batch size, learning rate and attention head number to 32, 1e-5 and 8 for both datasets. All experiments are carried out with an RTX 3090 GPU and the results are obtained by averaging 10 trials with different random seeds on testing data. 
Following \citet{Guo_Lee_Ulbricht_2020}, we train our model for 500 epochs due to two main reasons: (1)~Multi-task learning often needs more epochs to converge due to its complex objectives and task balancing. (2) We notice continued performance improvement beyond the usual training duration.

\begin{table*}[htb]
	\footnotesize
	\centering
	\begin{tabular}{l|ccc|ccc}
		\toprule
		\multirow{2}{*}{Methods}   & \multicolumn{3}{|c|}{\textit{Restaurant}}  & \multicolumn{3}{c}{\textit{Laptop}} \\
		\cmidrule{2-7}
		& Precision & Recall  & F1  & Precision & Recall  & F1 \\
		\midrule
		%Double-Propagation (DP)          & 0.3467 & 0.1508 & 0.2104 & 0.1304 & 0.0057 & 0.0800 \\
		%JET           & 0.5981 & 0.2894 & 0.3901 & 0.4452 & 0.1625 & 0.2381 \\
		TAS           & 0.2629 & 0.4629 & 0.3353 & 0.4715 & 0.1922 & 0.2731 \\
		Extract-Classify  & 0.3854 & 0.5296 & 0.4461 & 0.4556 & 0.2948 & 0.3580 \\
		{\sc Paraphrase} & 0.4362 & 0.3619 & 0.3956 & 0.3636 & 0.2963 & 0.3265\\
		GEN-NAT-SCL & 0.4893 & 0.4051 & 0.4432 & 0.3713 & 0.3244 & 0.3463\\
		BART-CRN & 0.5084 & 0.4710 & 0.4890 & 0.4816 & 0.3183 & 0.3832\\
		\midrule
		iACOS         & \textbf{0.5724} & \textbf{0.5321} & \textbf{0.5515} & \textbf{0.4959} & \textbf{0.3465} & \textbf{0.4080} \\
		std          & $\pm$0.0095 & $\pm$0.0079 & $\pm$0.0072 & $\pm$0.0121 & $\pm$0.0101 & $\pm$0.0082 \\
		%\midrule
		%iACOS' improvement $\uparrow$  &        &        & \textbf{23.63}\%&        &        & \textbf{12.07}\%\\
		\bottomrule
	\end{tabular}
	\caption{Performance comparison on the two datasets with implicit aspects and opinions. The results of compared methods are from the previous works~\cite{Cai_Xia_Yu_2021,Xiong_Yan_Wu_2023}.}
	\label{tab:f1-imp}
\end{table*}

\begin{table*}[htb]
	\footnotesize
	\centering
	\begin{tabular}{l|c@{~~~}c@{~~~}c@{~~~}c|c@{~~~}c@{~~~}c@{~~~}c}
		\toprule
		\multirow{2}{*}{Methods}   & \multicolumn{4}{|c|}{\textit{Restaurant}}  & \multicolumn{4}{c}{\textit{Laptop}} \\
		\cmidrule{2-9}
		& EA\&EO  & IA\&EO  & EA\&IO & IA\&IO  & EA\&EO  & IA\&EO  & EA\&IO & IA\&IO \\
		\midrule
		%DP            & 0.2602 & N/A & N/A & N/A & 0.0980 & N/A & N/A & N/A \\
		%JET           & \textbf{0.5230} & N/A & N/A & N/A & \textbf{0.3570} & N/A & N/A & N/A \\
		TAS           & 0.3360 & 0.3184 & 0.1403 & 0.3976 & 0.2610 & 0.4154 & 0.1090 & 0.2115\\
		Extract-Classify  & 0.4496 & 0.3466 & 0.2386 & 0.3370 & 0.3539 & 0.3900 & 0.1682 & 0.1858\\
		{\sc Paraphrase} & 0.3852 & 0.3780 & 0.1667 & 0.3850 & 0.3130 & 0.3892 & 0.2111 & 0.3556\\
		GEN-NAT-SCL & 0.4692 & 0.3053 & 0.2051 & 0.3763 & 0.3593 & 0.407 & 0.2085 & 0.3022\\
		BART-CRN & 0.5413 & \textbf{0.5064} & 0.1893 & 0.4286 & 0.3891 & 0.5430 & \textbf{0.2450} & 0.4071\\
		\midrule
		iACOS         & \textbf{0.6166} & 0.4778 & \textbf{0.2491} & \textbf{0.4345} & \textbf{0.4201} & \textbf{0.5808} & 0.2394 & \textbf{0.4124}\\
		std     & $\pm$0.0093 & $\pm$0.0139 & $\pm$0.0085 & $\pm$0.0108 & $\pm$0.0101 & $\pm$0.0078 &$\pm$0.0103 & $\pm$0.0105 \\
		%\midrule
		%$\uparrow$  &    17.90\% &  37.85\% &   4.40\% &    9.28\% &  17.70\% &   42.27\% &  26.22\% &  84.16\%\\
		\bottomrule
	\end{tabular}
	\caption{F1 score on testing subsets with different aspect \& opinion types. E, I, A and O denote Explicit, Implicit, Aspect and Opinion, respectively. The results of compared methods are from the previous works~\cite{Cai_Xia_Yu_2021,Xiong_Yan_Wu_2023}.}
	\label{tab:f1-4types}
\end{table*}

\begin{table*}[htb]
	\footnotesize
	\centering
	\begin{tabular}{l|c@{~~~}c@{~~~}c@{~~~}c|c@{~~~}c@{~~~}c@{~~~}c}
		\toprule
		\multirow{2}{*}{Methods}   & \multicolumn{4}{|c|}{\textit{Restaurant}}  & \multicolumn{4}{c}{\textit{Laptop}} \\
		\cmidrule{2-9}
		& EA\&EO  & IA\&EO  & EA\&IO & IA\&IO  & EA\&EO  & IA\&EO  & EA\&IO & IA\&IO \\
		\midrule
		iACOS         & 0.6166 & 0.4778 & 0.2491 & 0.4345 & 0.4201 & 0.5808 & 0.2394 & 0.4124\\
		Random	      & 0.5554 & 0.4508 & 0.2464 & 0.2715 &	0.3975 & 0.5452 & 0.1857 & 0.2190\\
		None	      & 0.4360 & 0.1751 & 0.1385 & 0.1931 & 0.3529 & 0.1992 & 0.1434 & 0.0817\\
		\bottomrule
	\end{tabular}
	\caption{Effect of negative samples on the extraction of different aspect \& opinion types.}
	\label{tab:nsm}
\end{table*}

\subsection{Experimental Results} \label{subsec:ra}

\textbf{Convergence analysis.}
Our iACOS shows quite stable and consistent performance on different trials. \figurename~\ref{fig:convergent} depicts the convergent process with respect to the number of epochs on a trial, in which 500 epochs are equally divided into five bins and the mean performance is calculated for each bin, along with standard deviation boundaries. After 200 epochs, iACOS reaches relatively stable performance and the F1 score steadily and slowly increases on both validation data and testing data. After 300 epochs, the standard deviation is negligible, and although the validation F1 score remains increasing, the F1 score records the maximum value at 400 epochs on \textit{Restaurant} testing data and at 500 epochs on \textit{Laptop} testing data. Hereafter, unless otherwise specified, we report results at 400 epochs on testing data.

\textbf{Overall comparison.}
\tablename~\ref{tab:f1-imp} compares the performance of all evaluated methods. Our iACOS consistently achieves the best results averaged at ten random trials with negligible standard deviation on both datasets. Note that the original references~\cite{Cai_Xia_Yu_2021,Xiong_Yan_Wu_2023} have not reported the standard deviation for the other methods. Compared to the second best BART-CRN, iACOS relatively improves the F1 score by 12.78\% and 6.47\% on \textit{Restaurant} and \textit{Laptop} datasets, respectively. 

Further, we conform to the approach outlined in the reference~\cite{Cai_Xia_Yu_2021} by focusing on the four principal combinations: EA\&EO, IA\&EO, EA\&IO, and IA\&IO. \tablename~\ref{tab:f1-4types} demonstrates the performance on different testing subsets. iACOS reaches the highest F1 score among all evaluated methods in most cases, especially on the two testing subsets: EA\&EO and IA\&IO. BART-CRN has a better performance on IA\&EO of \textit{Restaurant} and EA\&IO of \textit{Laptop}. One reasonable explanation
is that the proportion of IA\&EO in \textit{Restaurant} or EA\&IO in \textit{Laptop} is higher than the other implicit testing subsets as shown in \tablename~\ref{tab:data}, which helps BART-CRN to fully capture the input features~\cite{Xiong_Yan_Wu_2023}. These results indicate the effectiveness of iACOS with informative and adaptive negative samples.

%
%\begin{figure*}[!t]
%	\centering
%	\begin{minipage}{0.5\textwidth}
%		\centering
%		\subfloat[\textit{Restaurant}]{\label{subfig:rest16-nsm}%
%			\includegraphics[width=0.5\textwidth]{rest16-sampling-method-s}}
%		\hfill
%		\subfloat[\textit{Laptop}]{\label{subfig:laptop-nsm}%
%			\includegraphics[width=0.5\textwidth]{laptop-sampling-method-s}}
%		\caption{Effect of negative sampling methods}
%		\label{fig:nsm}
%	\end{minipage}\hfill
%	\begin{minipage}{0.5\textwidth}
%		\centering
%		\subfloat[\textit{Restaurant}]{\label{subfig:rest16-ablation}%
%			\includegraphics[width=0.5\textwidth]{rest16-ablation-study-s}}
%		\hfill
%		\subfloat[\textit{Laptop}]{\label{subfig:laptop-ablation}%
%			\includegraphics[width=0.5\textwidth]{laptop-ablation-study-s}}
%		\caption{Ablation experiment}
%		\label{fig:ablation}
%	\end{minipage}
%\end{figure*}

\begin{figure}[!t]
	\centering
	\subfloat[\textit{Restaurant}]{\label{subfig:rest16-nsm}%
		\includegraphics[width=0.5\columnwidth]{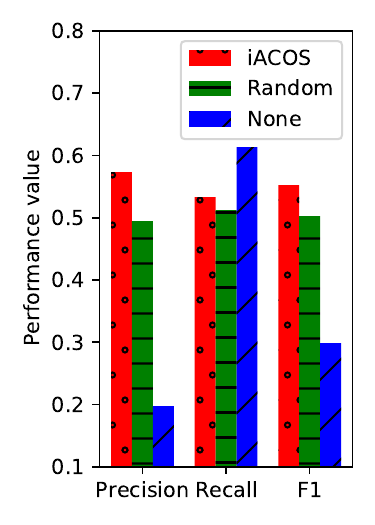}}
	\hfill
	\subfloat[\textit{Laptop}]{\label{subfig:laptop-nsm}%
		\includegraphics[width=0.5\columnwidth]{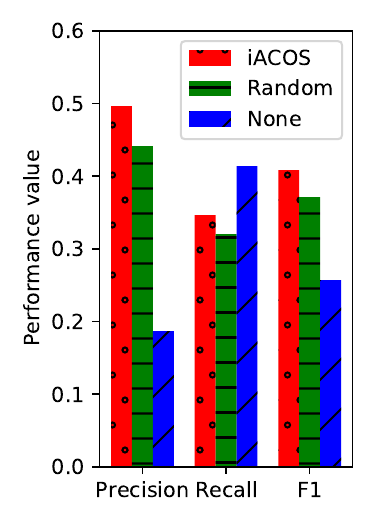}}
	\caption{Effect of negative samples}
	\label{fig:nsm}
\end{figure}

\textbf{Study on negative samples.}
\figurename~\ref{fig:nsm} depicts the effect of different negative sample construction methods with three findings. Firstly, ``None'' does not apply any negative samples, i.e., training with ground-truth quadruples $Y^+$ only. Its precision and F1 score decrease severely, even though it records the highest recall. One reason is that without negative samples, it is prone to extract more aspects and opinions from texts and results in proposing more quadruples. Secondly, the random method generates the sets of aspects and opinions randomly instead of employing the sequence labeling model presented in Section~\ref{subsec:co-ex}, and then follows the same remainder process as iACOS. The random method outperforms ``None'' in terms of the F1 scores on both datasets, which indicates that these negative samples are helpful to improve the model performance regardless of underlying sampling methods. Finally, iACOS constructs information and adaptive samples based on the aspect-opinion co-extraction results and increases the F1 score by 8\% at least on both datasets in comparison to the random method. This indicates a better construction method can bring larger performance improvement.  

Furthermore, \tablename~\ref{tab:nsm} shows the effect of negative samples on the extraction of different aspect \& opinion types. The ``None'' condition, which does not apply any negative samples, results in the lowest F1 scores across all cases. This observation allows us to conclude that negative samples have a positive effect on the extraction of all aspect and opinion types. This outcome occurs because our proposed method is not specifically tailored for IA\&IO.

\begin{figure}[!t]
	\centering
	\subfloat[\textit{Restaurant}]{\label{subfig:rest16-ablation}%
		\includegraphics[width=0.5\columnwidth]{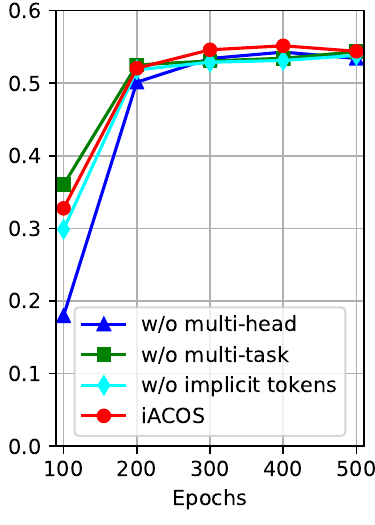}}
	\hfill
	\subfloat[\textit{Laptop}]{\label{subfig:laptop-ablation}%
		\includegraphics[width=0.5\columnwidth]{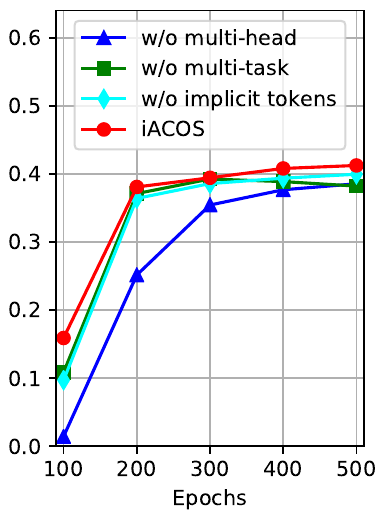}}
	\caption{Ablation experiments}
	\label{fig:ablation}
\end{figure}

\textbf{Ablation study.} 
\figurename~\ref{fig:ablation} illustrates the influence of implicit tokens, multi-head attention and multi-task learning in iACOS. (1)~Without adding implicit tokens using the [CLS] token of BERT for implicit aspects and opinions, the performance of iACOS degrades noticeably in both the \textit{Restaurant} and \textit{Laptop} domains. The reason is that this alternative method cannot differentiate between implicit aspects and implicit opinions, resulting in significantly lower performance on IA\&EO and EA\&IO, particularly on IA\&IO. (2)~Without multi-head attention by simply taking average of all vectors of ${\bf H}$ in \equationautorefname~(\ref{eq:pair-att}), iACOS encounters underfitting and reports the lowest F1 score, especially on \textit{Laptop} domain that has much more aspect categories than \textit{Restaurant} domain. This result justifies that multi-head attention plays an important role in augmenting model capacity by enabling the simultaneous capture of diverse features and relationships within the input data, leading to improved representation learning and overall performance gains. (3)~Without multi-task learning by minimizing the sum of $L_1$ and $L_2$ rather than the total loss in \equationautorefname~(\ref{eq:loss-total}), iACOS quickly converges, suffers from overfitting, and deteriorates performance on \textit{Laptop} domain. This result indicates that the multi-task learning enhances model generalization, improves predictive accuracy, and enables effective knowledge transfer across related tasks.
 

%% file: conclusion_naacl.tex
\section{Conclusion}\label{sec:conclusion}
In this paper, we propose iACOS, a novel approach for extracting implicit sentiment quadruples with multi-label classifier and multi-head attention over the context-aware representation of implicit aspects and opinions. Furthermore, we devise an informative and adaptive sample construction method for generating negative examples to train multiple classifiers by multi-task learning. Experiment results have verified our method's effectiveness and superiority in comparison to existing strong baselines.

\section*{Limitations}
%There are several research lines in our future work. First, we will explore theoretical justification for the proposed informative and adaptive sample construction method. Second, our model will be evaluated not only on the quadruple extraction task but also on other ABSA tasks. Third, we will investigate the effect of various hyper-parameters, such as batch size, learning rate, and attention head count. Our study provides valuable insights into the effectiveness of iACOS for extracting implicit sentiment quadruples and suggests areas for future research.

First, we have not provided theoretical justification for the proposed informative and adaptive sampling method. Second, our model is only evaluated on the quadruple extraction task and its effectiveness on other ABSA tasks is unknown. Third, we have not extensively investigated the effect of various hyper-parameters, e.g., the batch size, learning rate and attention head number. Lastly, we have not explored applying our negative sample construction method to large language models. Despite these limitations, our study provides valuable insights into the effectiveness of iACOS for extracting implicit sentiment quadruples and suggests areas for future research.